%% file: main.tex
\documentclass[10pt,twocolumn,letterpaper]{article}
\usepackage[pagenumbers]{cvpr} 
\input{preamble}

\definecolor{cvprblue}{rgb}{0.21,0.49,0.74}
\usepackage[pagebackref,breaklinks,colorlinks,allcolors=cvprblue]{hyperref}
\def\confName{CVPR}
\def\confYear{2027}
\hypersetup{
  pdftitle={TIVS: Mapping the Sensitivity Landscape of Visual Token Transformations in MLLMs},
  pdfauthor={Yikang Qiao, Jianye Zeng, Duan Huang, Ziying Song}
}

\title{TIVS: Mapping the Sensitivity Landscape of Visual Token Transformations in MLLMs}
\author{%
Yikang Qiao\textsuperscript{*}\textsuperscript{1}\qquad
Jianye Zeng\textsuperscript{*}\textsuperscript{2}\qquad
Duan Huang\textsuperscript{$\dagger$}\textsuperscript{1}\qquad
Ziying Song\textsuperscript{$\dagger$}\textsuperscript{3}\\
\textsuperscript{1}Central South University\qquad
\textsuperscript{2}University of Science and Technology of China\\
\textsuperscript{3}Yanshan University\\
{\small \textsuperscript{*}These authors contributed equally to this work.\quad
\textsuperscript{$\dagger$}Corresponding authors.}
}

\begin{document}
\maketitle
\begin{abstract}
Recent efforts to improve the efficiency of multimodal large language models (MLLMs) have explored visual token pruning, quantization, and feature compression.
However, the properties that determine whether transformed visual tokens remain usable by a frozen decoder are not well characterized.
We introduce \textbf{TIVS (Token Identity Versus Statistics)}, a training-free intervention framework that separately studies sequence permutation, token retention, numerical precision, and statistical replacement under controlled conditions.
On 192 POPE questions, three token permutations change TinyLLaVA accuracy by only $-0.17$ percentage points on average, while changing 4.69\% of individual predictions. Diagonal and rank-64 Gaussian replacements decrease accuracy by 25.0 and 28.1 points on the same subset.
Further experiments with exact moment constraints, quantization error, token retention, and additional visual tasks reveal how strongly the effects depend on the original feature values and the evaluation setting.
These findings provide an empirical basis for assessing visual token transformations in frozen MLLMs and designing more reliable multimodal interfaces.
\end{abstract}

\input{sections/01_introduction}
\input{sections/02_related_work}
\input{sections/03_framework}
\input{sections/04_experimental_setup}
\input{sections/05_results}
\input{sections/08_conclusion}

\input{sections/09_llm_statement}

{\small
\bibliographystyle{ieeenat_fullname}
\bibliography{references}
}
\end{document}

%% file: preamble.tex
\usepackage{float}
\usepackage{microtype}
\usepackage{mathtools}
\newcommand{\method}{\textsc{TIVS}}


\usepackage{stfloats}

%% file: sections/01_introduction.tex
\section{Introduction}
\label{sec:intro}

\begin{figure}[t]
    \centering
    \includegraphics[width=\columnwidth]{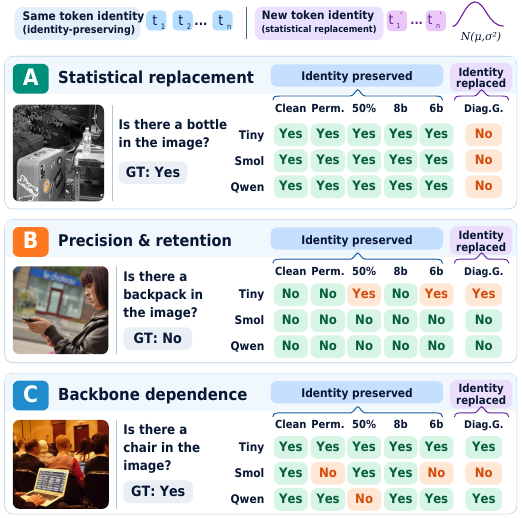}
    \caption{\textbf{Overview of TIVS.}
    TIVS systematically investigates visual token transformations
    in frozen MLLMs. The results highlight the importance of preserving
    token-level numerical identity for visual understanding.}
    \label{fig:overview1}
\end{figure}

Multimodal large language models (MLLMs) represent images as sequences of visual features consumed by language decoders. These sequences improve visual grounding but also increase inference cost. Recent methods have reduced visual token budgets through attention-based pruning, feature selection, and compression. FastV removes tokens using decoder attention \cite{chen2024fastv}. SparseVLM employs text-guided selection \cite{zhang2025sparsevlm}. VisionZip compresses visual encoder features \cite{yang2025visionzip}. Their results show that many visual tokens can be removed or combined while preserving useful model behavior. These compression strategies modify different aspects of visual representations. Token pruning changes the available token set, while quantization and feature transformations alter the numerical content of visual features.

However, the properties that determine whether transformed visual tokens remain usable by frozen MLLMs are not yet well understood. In particular, the relative importance of token-level numerical identity, sequence order, and aggregate feature statistics remains unclear. These properties represent different aspects of visual information. Token permutation changes the sequence order while preserving the original feature vectors. Quantization modifies their numerical values, whereas statistical replacement generates new vectors from the original feature distribution. Understanding how these transformations affect model predictions can provide insights into the information required by frozen visual interfaces.

A visual token sequence contains information at multiple levels. Individual token vectors encode specific numerical features, while their aggregate statistics describe the overall feature distribution. These properties are not equivalent. Two sequences may exhibit similar means and covariances while containing different feature vectors. Conversely, token permutation preserves the original vectors while changing their positions. Understanding how these differences affect downstream predictions is essential for identifying which properties of visual representations support reliable multimodal inference.

We introduce \method{} (\textbf{T}oken \textbf{I}dentity \textbf{V}ersus \textbf{S}tatistics), a training-free framework for controlled visual token interventions. TIVS fixes the model and evaluation examples, varies one intervention at a time, and measures both accuracy and individual prediction changes. We study permutations, token retention, feature quantization, and Gaussian replacements. We also test exact finite-sample moment constraints and perturbations matched by magnitude. These controls connect the properties of the transformed features to downstream predictions.

\begin{figure*}[t]
\centering
\includegraphics[width=.94\textwidth]{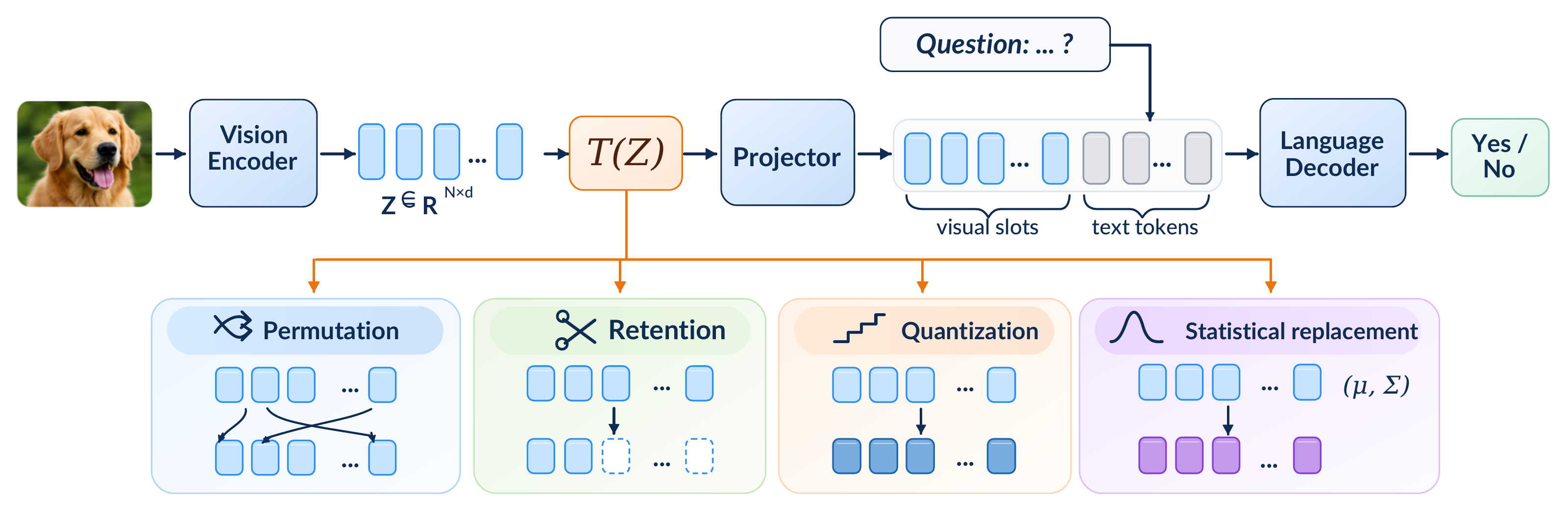}
\caption{\textbf{Controlled visual token interventions.} TinyLLaVA features are transformed before the frozen projector and inserted into the visual positions of the language sequence. Retention reduces the number of slots. Decoder and binary readout remain fixed.}
\label{fig:overview2}
\end{figure*}

Beyond aggregate accuracy, we examine how visual token transformations affect individual model decisions. Two configurations can achieve identical accuracy while producing different predictions on the same questions. We therefore evaluate prediction consistency and logit margins alongside conventional accuracy. This paired analysis provides a more detailed characterization of how visual token transformations influence the behavior of frozen decoders.

Our experiments reveal a clear difference between transformations that preserve the original token vectors and those that replace their numerical content. On 192 POPE questions \cite{li2023pope}, token permutation changes TinyLLaVA accuracy by only $-0.17$ percentage points on average, while diagonal and rank-64 Gaussian replacements reduce accuracy by 25.0 and 28.1 points, respectively. Despite the small change in aggregate accuracy, permutation still alters 4.69\% of individual predictions. Further experiments with exact covariance matching, numerical precision, token retention, and different model backbones provide additional evidence of how visual feature properties influence frozen MLLM behavior. Our main contributions are summarized as follows: 

\begin{itemize}
    \item We introduce TIVS, a training-free framework
    that systematically investigates the effects of
    token order, retention, numerical precision, and
    statistical properties on frozen MLLMs.

    \item We establish a paired evaluation protocol
    that jointly measures prediction accuracy,
    decision consistency, and logit margins,
    revealing changes in individual predictions
    that aggregate accuracy can conceal.

    \item We provide systematic experimental evidence
    across token transformations, feature layers,
    and model backbones, highlighting the importance
    of token-level numerical identity for reliable
    visual representations.
\end{itemize}

%% file: sections/02_related_work.tex
\section{Related Work}
\label{sec:related}

\paragraph{Visual token selection.}
Efficient MLLMs commonly reduce the number of visual tokens passed to the decoder. FastV prunes tokens using attention signals \cite{chen2024fastv}. ATP-LLaVA adapts pruning decisions across inputs and layers \cite{ye2025atpllava}. DivPrune emphasizes token diversity \cite{alvar2025divprune}. TopV considers inference compatibility \cite{yang2025topv}. VFlowOpt uses visual information flow to guide pruning \cite{yang2025vflowopt}. TransPrune analyzes representation transitions \cite{li2026transprune}. VisionZip compacts visual encoder outputs \cite{yang2025visionzip}. These methods motivate our study of retention budgets and selection strategies.

\paragraph{Feature compression and transformation.}
Other methods modify visual representations to create shorter inputs. LLaVA-PruMerge merges related visual features \cite{shang2025prumerge}. ApET uses approximation error in compression \cite{ma2026apet}. CORE consolidates object-centric information \cite{lei2026core}. Hybrid Token Compression combines discrete and continuous representations \cite{zhang2026htc}. EvoComp studies learned compression policies \cite{song2026evocomp}. While TIVS analyzes how numerical modification and statistical replacement affect a fixed visual interface.

\paragraph{Visual token representations.}
Recent analyses examine feature sparsity, redundancy, and token importance. Prior work studies sparsity and redundancy in MLLMs \cite{fan2026tokens}. Feather connects pruning with spatial coverage \cite{endo2025feather}. Skip-Vision identifies redundant computation across layers \cite{zeng2025skipvision}. Variation-aware dropping uses temporal feature changes \cite{chen2026v2drop}. Random-selection comparisons probe token importance \cite{wang2026pruning}. Rethinking Token Reduction investigates multimodal interactions during token reduction \cite{wang2026rethink}. Our controlled interventions focus on the distinction between original feature values, their sequence assignments, and their empirical moments.

%% file: sections/03_framework.tex
\section{TIVS: Controlled Visual Token Interventions}
\label{sec:method}
\subsection{Frozen visual interfaces}
Given an image $I$ and question $q$, let $Z\in\mathbb R^{N\times d}$ be the visual token matrix. In our main TinyLLaVA protocol, $Z$ contains the non-CLS features of the selected CLIP hidden state. Let $P$ denote the frozen projector, $D$ the frozen decoder, and $H(q)$ the text embeddings. For a transformation $T$, we define
\begin{align}
Z_T &= T(Z)\in\mathbb R^{N_T\times d},\label{eq:ztrans}\\
U_T &= P(Z_T),\qquad
L_T=D\big(\mathrm{Inject}(H(q),U_T)\big).\label{eq:flow}
\end{align}
The number of image slots changes to $N_T$ under token removal. The prompt and model weights remain fixed. A second protocol applies comparable interventions at the \emph{decoder input} to support a matched cross-backbone comparison. The two injection sites are analyzed separately.

For POPE, let $\mathcal V_y$ and $\mathcal V_n$ denote the recorded valid single-token forms of yes and no. The binary margin and prediction are
\begin{align}
m_T&=\max_{v\in\mathcal V_y} L_T[v]
-\max_{v\in\mathcal V_n} L_T[v],\label{eq:margin}\\
\hat y_T&=\mathbf 1[m_T>0].\label{eq:decision}
\end{align}
A zero margin is scored as no. Other benchmarks use their task-specific readout as detailed in \Cref{sec:setup}.

\subsection{Numerical identity and statistical constraints}
We distinguish the original set of image-conditioned vectors from its sequence positions and aggregate statistics. The empirical mean and sample covariance are
\begin{align}
\mu(Z)&=\frac{1}{N}\sum_{i=1}^{N}z_i,\label{eq:mu}\\
\Sigma(Z)&=\frac{1}{N-1}\sum_{i=1}^{N}(z_i-\mu)(z_i-\mu)^\top.\label{eq:cov}
\end{align}
These moments describe the feature population. They do not uniquely determine its individual vectors. To characterize token identity independently of sequence order, we define the multiset of visual feature vectors as

\begin{equation}
\mathcal{M}(Z) =
\{\!\{z_1,z_2,\ldots,z_N\}\!\}.
\label{eq:identity}
\end{equation}

A permutation preserves this multiset exactly, while changing the assignment of feature vectors to sequence positions. In contrast, statistical constraints on the mean and covariance do not uniquely determine $\mathcal{M}(Z)$. This distinction allows TIVS to examine the roles of individual feature values, sequence order, and population statistics through controlled interventions.

\paragraph{Permutation and retention.}
For a permutation matrix $\Pi\in\{0,1\}^{N\times N}$,
\begin{equation}
T_{\mathrm{perm}}(Z)=\Pi Z,\quad
\mu(\Pi Z)=\mu(Z),\quad \Sigma(\Pi Z)=\Sigma(Z).
\label{eq:perm}
\end{equation}
Permutation preserves the feature multiset exactly. Token retention selects $K$ rows using $S_K\in\{0,1\}^{K\times N}$:
\begin{equation}
T_{\mathrm{keep}}(Z)=S_KZ,\qquad K=\lfloor \rho N\rfloor,
\label{eq:retention}
\end{equation}
where $\rho\in\{0.25,0.50,0.75\}$ in the new retention sweeps. The selected rows may come from the first positions, last positions, uniform spacing, or randomized subsets. The original relative order of retained rows is preserved.

\paragraph{Affine min--max quantization.}
For each image-level tensor, let $a=\min(Z)$, $b=\max(Z)$ and $B$ be the requested bit width. The implemented round-and-reconstruct operation is
\begin{align}
\delta_B&=\frac{\max(b-a,10^{-6})}{2^B-1},\\
Q_B(Z)&=a+\delta_B\,\mathrm{clip}\!\left(
\mathrm{round}\!\left(\frac{Z-a}{\delta_B}\right),0,2^B-1\right).
\label{eq:quant}
\end{align}
It uses $2^B$ affine grid points and ties-to-even rounding. For in-range scalar entries, the ideal FP32 rounding error is bounded by $\delta_B/2$. The features are then cast to the model dtype. This experiment probes numerical precision at the frozen interface.

\paragraph{Moment-based replacements.}
Let $\sigma_j^2=\Sigma(Z)_{jj}$. A diagonal Gaussian surrogate is
\begin{equation}
[T_{\mathrm{diag}}(Z)]_i=\mu(Z)+\mathrm{diag}(\sigma)\varepsilon_i,
\quad\varepsilon_i\sim\mathcal N(0,I_d).
\label{eq:diag}
\end{equation}
A rank-$r$ Gaussian surrogate uses the leading covariance factors:
\begin{equation}
[T_{\mathrm{pca}}(Z)]_i=\mu(Z)+U_r\Lambda_r^{1/2}\eta_i,
\quad\eta_i\sim\mathcal N(0,I_r),\ r=64.
\label{eq:pca}
\end{equation}
These Gaussian families target moments in expectation. We additionally test exact finite-sample constraints. For an orthogonal row-mixing matrix $R$ with $R\mathbf 1=\mathbf 1$, define
\begin{equation}
Z' =\mathbf 1\mu(Z)^\top +R\big(Z-\mathbf 1\mu(Z)^\top\big).
\label{eq:rowmix}
\end{equation}
Let $Z_c = Z-\mathbf{1}\mu(Z)^\top$ denote the centered feature matrix. Since $R$ is orthogonal and satisfies $R\mathbf{1}=\mathbf{1}$, the transformed features preserve the empirical mean and covariance:

\begin{align}
\mu(Z') &= \mu(Z),\\
\Sigma(Z')
&= \frac{1}{N-1}Z_c^\top R^\top R Z_c\\
&= \Sigma(Z).
\label{eq:exact_cov}
\end{align}

Despite preserving these statistics, row mixing changes the individual feature vectors in general. This provides a controlled setting for examining whether second-order statistical information is sufficient to maintain the behavior of a frozen decoder. The strong controls also include exact per-channel moments, rank-64 matching, per-token norm preservation, and displacement-matched Gaussian noise.

\begin{figure*}[t]
\centering
\includegraphics[width=.91\textwidth]{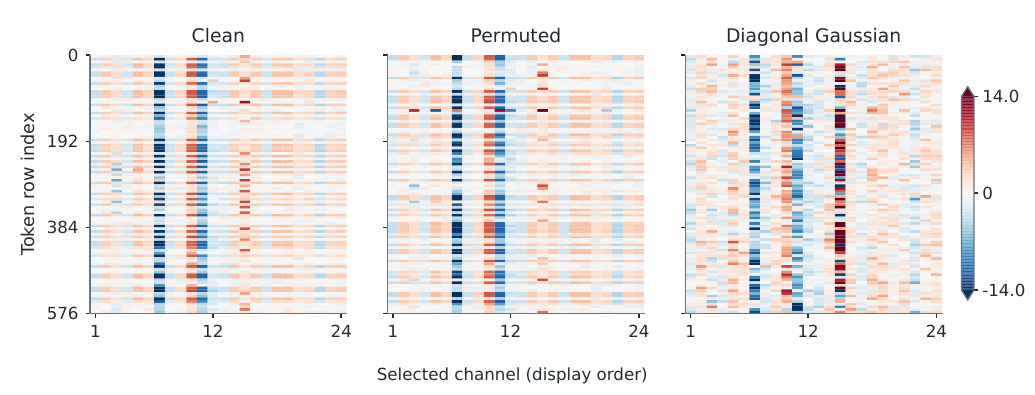}
\caption{\textbf{Measured token features.} Actual non-CLS TinyLLaVA features from one fixed image under clean, permutation and diagonal Gaussian replacement. All panels use the same 24 clean selected channels and token-row selection.}
\label{fig:features}
\end{figure*}

\subsection{Paired decisions and conditional stability}
For $M$ matched questions with labels $y_j$, define accuracy and prediction flip rate:
\begin{align}
A(T)&=\frac{1}{M}\sum_j\mathbf1[\hat y_{T,j}=y_j],\label{eq:acc}\\
F(T)&=\frac{1}{M}\sum_j\mathbf1[\hat y_{T,j}\ne\hat y_{0,j}],\label{eq:flip}\\
\Delta_A(T)&=100[A(T)-A(0)].\label{eq:delta}
\end{align}
Let $n_{\mathrm{cw}}$ and $n_{\mathrm{wc}}$ count clean-correct to wrong and clean-wrong to correct transitions. Then
\begin{equation}
M[A(T)-A(0)]=n_{\mathrm{wc}}-n_{\mathrm{cw}}.
\label{eq:transitions}
\end{equation}
This identity shows why an unchanged aggregate accuracy can coexist with altered predictions. We also inspect feature reconstruction error $\|T(Z)-Z\|_F/\|Z\|_F$ and paired logit margins. For binary question answering, every prediction flip changes the correctness of the corresponding answer. The flip rate can therefore be expressed as

\begin{equation}
M F(T) = n_{\mathrm{cw}} + n_{\mathrm{wc}}.
\label{eq:flip_decomposition}
\end{equation}

Together with \Cref{eq:transitions}, this shows that unchanged accuracy occurs when the numbers of corrected and degraded predictions are equal. A transformation can therefore preserve aggregate accuracy while changing individual decisions.

For aligned, equal-length features, assume the clean margin function is locally $L_m$-Lipschitz in the Frobenius norm. Then
\begin{equation}
|m_T-m_0|\le L_m\|T(Z)-Z\|_F.
\label{eq:lipschitz}
\end{equation}
The condition $|m_0|>L_m\|T(Z)-Z\|_F$ suffices to retain its sign under this assumption. TIVS measures margins and perturbations directly. A universal $L_m$ is not estimated, so we use \Cref{eq:lipschitz} as a conditional interpretation.

%% file: sections/04_experimental_setup.tex
\section{Experimental Setup}
\label{sec:setup}
\paragraph{Models and intervention locations.}
The principal frozen model is from \emph{TinyLLaVA: A Framework of Small-scale Large Multimodal Models} \cite{zhou2024tinyllava}, using the recorded 1.1B language backbone. The current local experiments use snapshot and non-CLS features from CLIP before the trained projector. An independent matched comparison applies transformations at decoder input to this model and SmolVLM-256M-Instruct. Historical Qwen2.5-VL-3B results \cite{bai2025qwen25vl} are retained as a separate protocol. All models are frozen.

\paragraph{Datasets.}
We evaluate POPE \cite{li2023pope} with a 64-question pilot and a new set of 192 questions from 32 distinct images. The larger set excludes the 11 pilot images. Shared 48-question subsets support matched interventions, cross-backbone comparisons and layer ablations. We also test 48 ChartQA questions and 48 MMBench English dev questions. ChartQA applies greedy short answer generation and 5\% relaxed numeric accuracy. MMBench scores the unrotated choice letters using candidate logits.

\paragraph{Evaluation and uncertainty.}
POPE uses the maximum-logit yes/no readout of \Cref{eq:margin}. We report correct counts, accuracy, differences from the clean arm and prediction flips. New paired 95\% intervals are computed with 10,000 image cluster bootstrap resamples. Multiple seeds are averaged per matched question before resampling. The intervals are exploratory and are not corrected for multiple comparisons. Image-cluster resampling keeps all questions associated with an image together. The paired comparisons retain within-image dependence across bootstrap draws. Historical aggregates are reported separately when item identities or logits are missing.

%% file: sections/05_results.tex
\section{Experimental Results and Analysis}
\label{sec:results}
\begin{figure}[!t]
\centering
\includegraphics[width=\columnwidth]{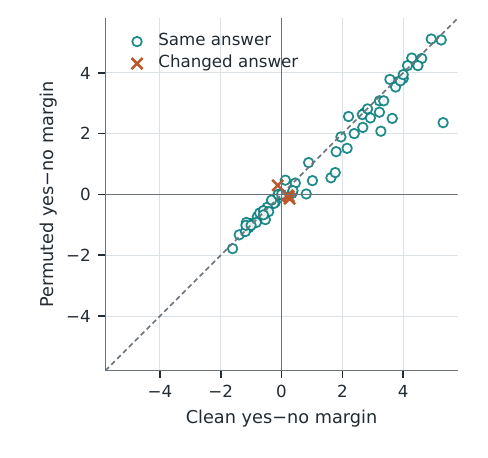}
\caption{\textbf{Paired binary margins under permutation.} The 64-question POPE run records four prediction flips at identical aggregate accuracy for clean and permuted tokens. Points represent actual recorded margins.}
\label{fig:margins}
\end{figure}

\begin{figure}[!htbp]
\centering
\includegraphics[width=\columnwidth]{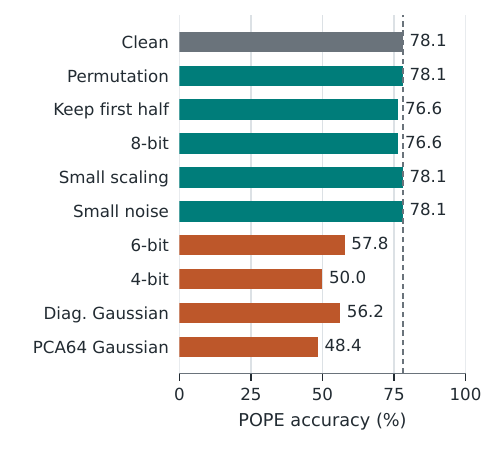}
\caption{\textbf{Historical 64-question POPE pilot.} Recorded accuracy across the original intervention arms. The dashed horizontal line marks 50/64 correct for clean.}
\label{fig:historicalbar}
\end{figure}

\begin{figure}[!htbp]
\centering
\includegraphics[width=\columnwidth]{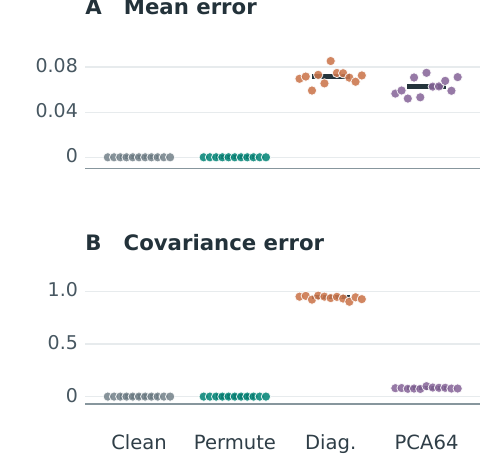}
\caption{\textbf{Finite-sample moment diagnostics.} Relative mean and covariance errors. Each point represents an individual image. Horizontal bars indicate median values.}
\label{fig:moment}
\end{figure}

\begin{figure*}[t]
\centering
\includegraphics[width=.91\textwidth]{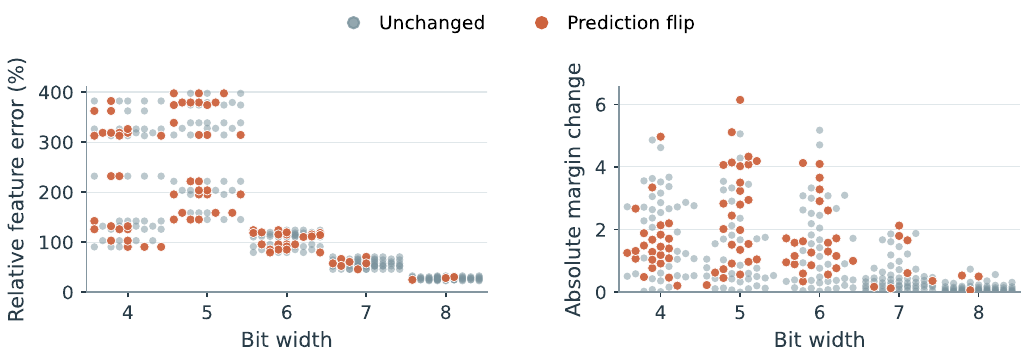}
\caption{\textbf{Quantization error and decision changes.} Each point is an observed question and bit width in the new 64-question run. Error is the relative Frobenius deviation before FP16 conversion.}
\label{fig:error}
\end{figure*}

\begin{figure}[!htbp]
\centering
\includegraphics[width=\columnwidth]{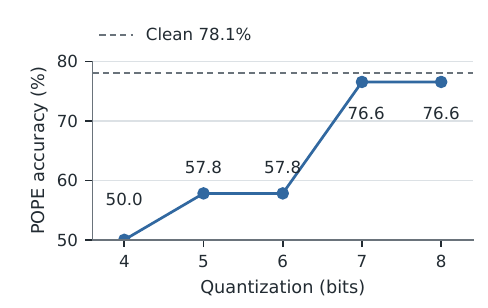}
\caption{\textbf{Bit-width sensitivity analysis.} POPE accuracy under quantization on 64 questions.}
\label{fig:bit}
\end{figure}

\begin{figure}[!htbp]
\centering
\includegraphics[width=.91\linewidth]{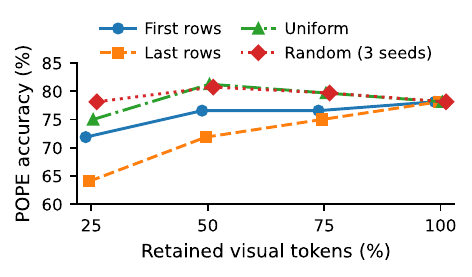}
\caption{\textbf{Token retention across budgets.} Measured first, last, uniform and random-subset retention on 64 POPE questions. The clean score is the 100\% reference.}
\label{fig:retention}
\end{figure}

\subsection{The main POPE intervention contrast}
The expanded POPE experiment contains 192 questions from 32 images. The clean TinyLLaVA model answers 152 questions correctly (79.17\%). Across three random permutations, mean accuracy changes by $-0.17$ percentage points with a paired image-cluster 95\% interval of $[-3.47,3.30]$ points. Prediction flips still occur on 4.69\% of questions on average. On the same set, diagonal Gaussian replacement yields 54.17\% accuracy and rank-64 Gaussian replacement yields 51.04\%. Both alter a much larger fraction of binary decisions (\Cref{tab:pope192}).

The results show a pronounced contrast between rearranging original image-conditioned features and replacing their joint values. The permutation effect is small in aggregate, although some individual answers differ. The real-valued feature representation therefore remains important even when the decoder tolerates extensive changes in feature order on this task.

\subsection{Prediction Consistency under Token Transformations}
To further investigate prediction consistency, we compare the model's predictions before and after visual token transformations. On 64 POPE questions, both the original and permuted tokens achieve 78.1\% accuracy, but their predictions differ on four questions. Additional permutation trials show similar behavior, with only minor changes in overall accuracy but differences in individual predictions. These results indicate that stable accuracy does not necessarily imply consistent predictions. For the original 64-question comparison, the four prediction flips occur despite identical clean and permuted accuracy. Under binary scoring, this means
that two originally correct predictions become incorrect, while two incorrect predictions become correct. The resulting changes cancel in aggregate accuracy. This observation illustrates why paired prediction analysis provides information beyond conventional accuracy comparisons.

\begin{table}[t]
\centering
\caption{\textbf{Expanded TinyLLaVA POPE evaluation.} All conditions share 192 questions and 32 images. Permutation is the average of three seeds. The 95\% intervals are paired image-cluster intervals for accuracy change.}
\label{tab:pope192}
\setlength{\tabcolsep}{3.2pt}\small
\begin{tabular}{lrrr}
\toprule
Intervention & Acc. (\%) & $\Delta$ (pt) & Flip (\%)\\
\midrule
Clean & 79.17 & 0.00 & ---\\
Permutation & 78.99 & $-0.17$ & 4.69\\
8-bit affine quant. & 78.65 & $-0.52$ & 3.65\\
6-bit affine quant. & 65.62 & $-13.54$ & 33.33\\
4-bit affine quant. & 48.96 & $-30.21$ & 50.00\\
Diagonal Gaussian & 54.17 & $-25.00$ & 41.67\\
Rank-64 Gaussian & 51.04 & $-28.12$ & 43.75\\
\bottomrule
\end{tabular}
\end{table}

\subsection{Impact of Statistical Transformations}
We probe whether preserving observed moments can recover frozen decoder predictions. On a shared 48 question POPE subset, the clean model achieves 42/48 (87.5\%). Exact per-channel moment matching achieves 52.08\% mean accuracy over three seeds. Exact mean and full covariance preservation through row mixing yields 47.92\%. A same-row norm control reaches 46.53\%, while a perturbation matched to the diagonal surrogate's Frobenius displacement reaches 58.33\%. \Cref{tab:strong} reports the measurements. The comparison between exact covariance matching and displacement-matched noise provides further insight into the role of feature statistics. Exact covariance matching achieves 47.92\% accuracy, while displacement-matched noise achieves 58.33\%. These interventions control different properties of the transformed features. Their distinct outcomes suggest that statistical agreement and perturbation magnitude capture different aspects of prediction sensitivity.

\begin{table}[!tbp]
\centering
\caption{\textbf{Strong controls on the shared 48-question POPE subset.} Means across three random seeds. Exact constraints refer to the FP32 intervention tensors before model-dtype conversion.}
\label{tab:strong}
\setlength{\tabcolsep}{3.1pt}\small
\begin{tabular*}{\columnwidth}{@{\extracolsep{\fill}}lrr}
\toprule
Intervention & Acc. (\%) & $\Delta$ (pt)\\
\midrule
Clean & 87.50 & ---\\
Permutation & 82.64 & $-4.86$\\
Exact per-channel moments & 52.08 & $-35.42$\\
Exact full covariance & 47.92 & $-39.58$\\
Exact rank-64 covariance & 50.69 & $-36.81$\\
Same row norms & 46.53 & $-40.97$\\
Matched-displacement noise & 58.33 & $-29.17$\\
\bottomrule
\end{tabular*}
\end{table}

The full covariance condition matches the clean empirical covariance to numerical precision before FP16 conversion. Its low accuracy strengthens the evidence that second-order population statistics are insufficient to reconstruct the original feature sequence for the frozen decoder. Matched displacement noise also changes decisions. The joint results identify feature-value alignment and intervention scale as important components of this behavior. The moment controls preserve aggregate channel statistics. They change the feature vectors occupying individual visual slots. The accuracy drop highlights the role of the image-specific token configuration.

\subsection{Task and backbone dependence}
\begin{figure*}[!t]
\centering
\includegraphics[width=.90\textwidth]{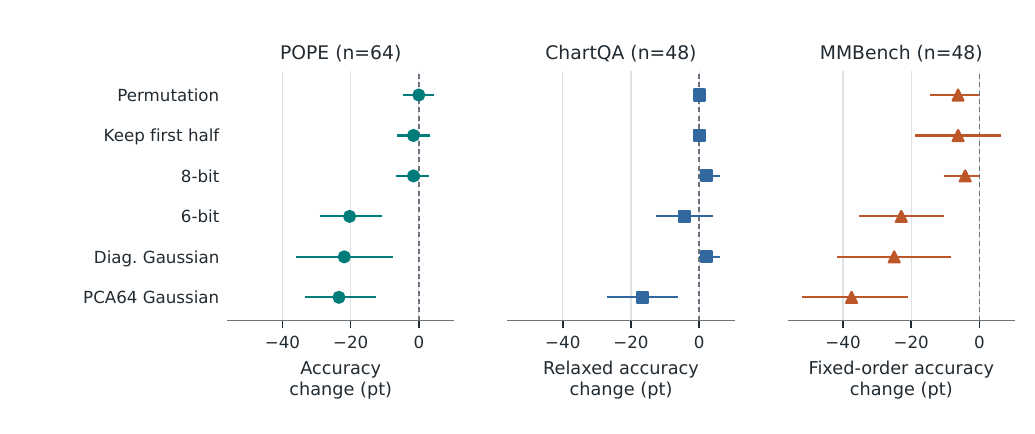}
\caption{\textbf{Sensitivity across tasks.} Changes relative to the task specific clean score, with exploratory paired image-cluster intervals where estimable. ChartQA uses short answer numeric tolerance. MMBench uses fixed option order.}
\label{fig:cross}
\end{figure*}
The cross-backbone experiment transforms visual embeddings at decoder input. TinyLLaVA, SmolVLM, and Qwen2.5-VL retain their native visual pipelines. The resulting accuracy changes measure the sensitivity of each complete frozen interface.

\subsection{Precision, margins, and token retention}
The local 64-question precision sweep adds 5 and 7 bits to the historical 4-, 6-, and 8-bit comparison. The resulting correct counts are 32, 37, 37, 49, and 49 for 4 through 8 bits. The clean model scores 50/64. The expanded 192-question run shows a similar ordering, with 94/192 for 4 bits, 126/192 for 6 bits, and 151/192 for 8 bits. The measured pattern depends on the affine min--max reconstruction of \Cref{eq:quant}.

The accuracy decline is concentrated in the 6--8-bit range under the affine min--max quantizer of \Cref{eq:quant}. The error measurements in \Cref{fig:error} expose the question-level changes accompanying this transition.

Retention sweeps evaluate 25\%, 50\%, and 75\% budgets. At a 50\% budget, keeping the first half yields 49/64, keeping the last half yields 46/64, and uniform spacing yields 52/64. Random subsets yield 54/64, 52/64, and 49/64 across three seeds. This variation shows the importance of the selected subset as well as the number of tokens retained.

Uniform selection spreads retained tokens across the sequence. First-half and last-half selection keep contiguous blocks. Their accuracy gap at the same 50\% budget shows a clear selection-pattern effect.

\begin{figure}[!t]
\centering
\includegraphics[width=\columnwidth]{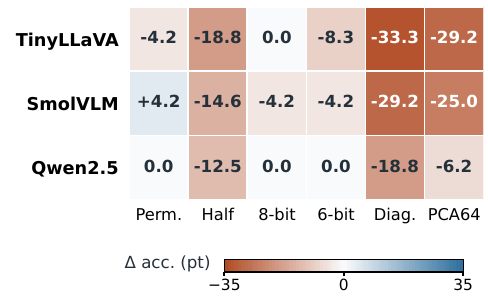}
\caption{\textbf{Matched decoder input backbone comparison.} Changes on the same 48 POPE questions. Every measured model uses its native visual encoder and prompt format.}
\label{fig:backbone}
\end{figure}

\Cref{tab:cross} extends the study beyond POPE using two 48-question task subsets. On MMBench English dev with fixed original option order, clean accuracy is 32/48 (66.7\%) and permutation reaches 29/48 (60.4\%). Eight-bit quantization reaches 30/48 (62.5\%). On ChartQA, clean and permutation both answer 8/48 correctly. Its low clean score makes the numerical comparison less informative for generalization. Different task readouts yield different sensitivity profiles.

\begin{table}[!tbp]
\centering
\caption{\textbf{Exploratory cross-task results on TinyLLaVA.} Each task has 48 questions, with its own scoring protocol.}
\label{tab:cross}
\setlength{\tabcolsep}{3.2pt}\small
\begin{tabular*}{\columnwidth}{@{\extracolsep{\fill}}lrrr}
\toprule
Task & Clean & Perm. & 8-bit\\
\midrule
POPE$^\dagger$ & 87.5 & 83.3 & 83.3\\
ChartQA & 16.7 & 16.7 & 18.8\\
MMBench & 66.7 & 60.4 & 62.5\\
\bottomrule
\end{tabular*}\par\vspace{2pt}{\scriptsize $^\dagger$Shared 48-question POPE subset.}
\end{table}

On the matched 48-question decoder-input subset, TinyLLaVA, SmolVLM-256M, and Qwen2.5-VL-3B achieve clean accuracies of 42/48, 39/48, and 45/48, respectively. Permutation changes accuracy by $-4.2$, $+4.2$, and $0.0$ percentage points across the three models. For Qwen2.5-VL-3B, retaining the first half of visual tokens reduces accuracy by 12.5 points, while 8-bit and 6-bit quantization preserve its clean accuracy on this subset. Diagonal Gaussian replacement decreases accuracy by 33.3, 29.2, and 18.8 points, respectively.

These results reveal backbone-dependent sensitivity to visual token transformations. Permutation produces different accuracy changes across models, whereas diagonal Gaussian replacement causes substantial degradation in all three backbones. PCA64 Gaussian replacement also reduces accuracy, but its effect varies across models. Together, these observations suggest that the importance of visual feature properties depends on the frozen model interface and the transformation applied.

\subsection{Layer-wise Sensitivity Analysis}
We further investigate the influence of visual feature depth on model performance. Using a shared subset of 48 POPE questions, we evaluate visual features extracted from four encoder layers with the same frozen projector. The clean accuracies are 58.3\%, 81.3\%, 87.5\%, and 68.8\% for layers $-8$, $-4$, $-2$, and $-1$, respectively. These results show that model performance is sensitive to the selected feature layer. We further analyze the effects of token permutation and quantization relative to the clean accuracy of each layer.

\begin{figure}[!t]
\centering
\includegraphics[width=\columnwidth]{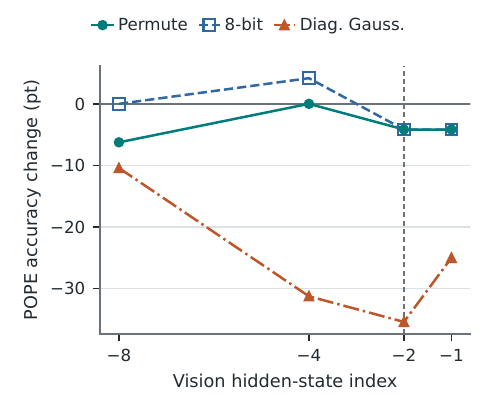}
\caption{\textbf{Layer sensitivity.} Differences relative to clean at each hidden state index, with the native $-2$ choice marked. Every layer is passed through the original frozen projector.}
\label{fig:layers}
\end{figure}

The native layer $-2$ gives the highest clean accuracy among the four tested depths. The frozen projector processes every layer with the same mapping. This exposes sensitivity at the encoder--projector interface. \Cref{fig:layers} measures each intervention against the clean accuracy at its extraction depth.

Across the four extraction depths, permutation changes accuracy by at most 6.3 points. Eight-bit quantization stays within 4.2 points of each clean baseline. Diagonal Gaussian replacement yields larger reductions at every depth, with the largest drop of 35.4 points at layer $-2$.

At layer $-4$, 8-bit quantization increases accuracy by 4.2 points. The other three layers show zero or negative changes.

%% file: sections/08_conclusion.tex
\section{Conclusion}
\label{sec:conclusion}
We introduced \method{}, a controlled study of visual token transformations in frozen MLLMs. Expanded POPE experiments, paired margins, and exact statistical controls distinguish the effects of changing token order from changing numerical feature content. Precision and retention sweeps show how accuracy depends on the magnitude and structure of a transformation. Cross-task and backbone comparisons further define the conditions under which these observations hold. TIVS provides reproducible evidence and practical measurements for designing reliable visual token interfaces. We also plan to leverage these findings to develop adaptive visual token compression methods in future work.

%% file: sections/09_llm_statement.tex
\section*{LLM Usage Statement}
Large language models were used for language editing. Concretely, we used GPT-6-Astra and DeepSeek-V4.1-Flash on the abstract, the introduction and other writing revision tasks to improve grammar, clarity and concision. Every AI assisted passage was reviewed and edited by the authors, who take full responsibility for the content of the paper.